\documentclass{article}

\usepackage{arxiv}

\usepackage[utf8]{inputenc} 
\usepackage[T1]{fontenc}    
\usepackage{hyperref}       
\usepackage{url}            
\usepackage{booktabs}       
\usepackage{amsfonts}       
\usepackage{nicefrac}       
\usepackage{microtype}      
\usepackage{lipsum}		
\usepackage{graphicx}
\usepackage[square,sort,comma,numbers]{natbib}
\usepackage{doi}

\usepackage{csquotes}

\title{Animal Inspired Application of a Variant of Mel Spectrogram for Seismic Data Processing}

\date{} 					
 \author{ \href{https://orcid.org/0000-0002-2725-5367}{\includegraphics[scale=0.06]{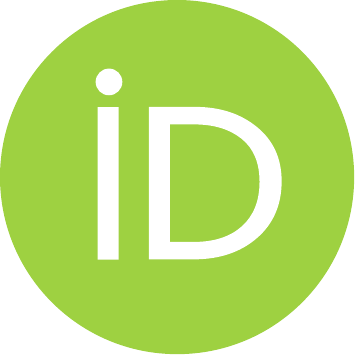}\hspace{1mm}Samayan Bhattacharya} \\
	Department of Computer Science and Engineering\\
	Jadavpur University \\
	 188, Raja Subodh Chandra Mallick Rd,\\ Jadavpur University Campus Area,\\ Jadavpur, Kolkata,\\ West Bengal 700032, India \\
	\texttt{samayan.bhattacharya@gmail.com} \\
	\And
	\href{https://orcid.org/0000-0001-7480-427X}{\includegraphics[scale=0.06]{orcid.pdf}\hspace{1mm}Sk Shahnawaz} \\
	Department of Computer Science and Engineering\\
	Jadavpur University \\
	 188, Raja Subodh Chandra Mallick Rd,\\ Jadavpur University Campus Area,\\ Jadavpur, Kolkata,\\ West Bengal 700032, India \\
	\texttt{skshahnawaz2909@gmail.com} \\
}

\renewcommand{\shorttitle}{\textit{Animal Inspired Application of a variant of Mel Spectrogram for Seismic Data Processing}}

\hypersetup{
pdftitle={Animal Inspired Application of a Variant of Mel Spectrogram for Seismic Data Processing},
pdfauthor={Samayan Bhattacharya, Sk Shahnawaz},
pdfkeywords={Mel spectrogram, Computer vision, seismic monitoring stations},
}

\begin{document}
\maketitle
\begin{displayquote}
\emph{"He did not notice the faint rumble that shook the earth"} - The Tiger in the Tunnel by Ruskin Bond\\
\end{displayquote}

\begin{abstract}
	Predicting disaster events from seismic data is of paramount importance and can save thousands of lives, especially in earthquake-prone areas and habitations around volcanic craters. The drastic rise in the number of seismic monitoring stations in recent years has allowed the collection of a huge quantity of data, outpacing the capacity of seismologists. Due to the complex nature of the seismological data, it is often difficult for seismologists to detect subtle patterns with major implications. Machine learning algorithms have been demonstrated to be effective in classification and prediction tasks for seismic data. It has been widely known that some animals can sense disasters like earthquakes from seismic signals well before the disaster strikes. Mel spectrogram has been widely used for speech recognition as it scales the actual frequencies according to human hearing. In this paper, we propose a variant of the Mel spectrogram to scale the raw frequencies of seismic data to the hearing of such animals that can sense disasters from seismic signals. We are using a Computer vision algorithm along with clustering that allows for the classification of unlabelled seismic data.
\end{abstract}

\keywords{Mel spectrogram \and Computer vision \and seismic monitoring stations}

\section{Introduction}
Understanding seismological data is important for disaster prediction as well as understanding the physical properties of the earth’s crust. In the past years, the drastic rise in the number of seismic monitoring stations has transformed the field of seismological research from an observation-based to a data-driven science \cite{1}. In general, earthquake seismology problems fall into three categories: probabilistic risk assessment \cite{2,3,4,5,6}, earthquake recognition for data mining and early-earthquake detection \cite{7,8,9,10,11}, and earthquake prediction for a warning system \cite{12,13}

Seismology studies are conducted with four objectives: (1) disaster preparation or adjustment,  (2) actions to reduce long-term the disaster of earthquake,  (3) disaster response strategies, and (4) post-disaster recovery planning, which are known as the strategies of preparedness, mitigation, response, and recovery, respectively. The challenges in the detection of every earthquake from seismic data are:  (1) a huge amount of noisy data in the seismic records and (2) many earthquake events are undetected\cite{11}. As a case in point, the nonvolcanic tremors were first observed in southwestern Japan only two decades ago\cite{14} This is because the weak signal generated by these tremors is hard to detect in certain regions. Hence, efficient detection of seismic signals would allow us to better understand the processes of seismic zones and hence predict earthquakes effectively.
In recent years Machine Learning algorithms have been demonstrated to be effective in classification and prediction tasks. Unlike other AI algorithms where the features have to be manually defined\cite{15,16,17} machine learning enables the computer to select relevant features on its own, it does require huge amounts of labeled data to train successfully. This requires a lot of man-hours and is a major disadvantage. A better alternative is the use of unsupervised learning, like clustering algorithms, that allow training on unlabelled data. The computer can separate the data into distinct classes based on similarities among members of a particular class. Unsupervised learning has been applied to the data from volcano monitoring systems\cite{18,19,20,21}, induced seismicity\cite{22,23}, global seismicity\cite{9}, and local vs distance earthquakes\cite{24}.

\section{Proposed method}
Mel spectrograms are widely used in speech recognition. The raw frequency data is converted to mel scale using the following formula:
\begin{equation}
    f’= 2595.log(1+(f/700))
\end{equation}
The rationale behind this is, it has been empirically proven that humans do not perceive frequencies linearly but logarithmically\cite{25}

In this paper, we convert the raw seismological data using the formula
\begin{equation}
	f’=c1.log(1+(f/c2))
\end{equation}
where, c1 and c2 are constants, derived empirically.

This is based on the fact that some animals can predict earthquakes from seismic signals\cite{26}. Therefore, it must help to adjust the spectrogram from human hearing to the hearing of those animals. From a machine learning point of view, this operation reduces the variance in data, unimportant for the given task, and increases variance for the important data. Thus, it makes it easier for the machine learning model to learn the important features for the prediction or classification task.

The obtained frequencies are converted to a spectrogram using a short-time Fourier transform and passed to the computer vision network.

In this experiment, we use a Convolutional Neural Network (CNN)\cite{27} to extract features from the spectrogram. The CNN is a form of deep neural network that uses multiple filters with trainable weights to extract features, usually from a 2D representation of the data. The CNN we used is a resnet\cite{28}. Resnet consists of layers of convolutions with skip connections between some layers, along with activation layers (we used ReLU) and pooling layers. Each layer of convolution applies a trainable filter to the data passed to it. We used a combination of max and average pooling layers in our model. The output of the resnet was fed into a clustering model.

\begin{figure}
    \centering
    \includegraphics[scale=0.5]{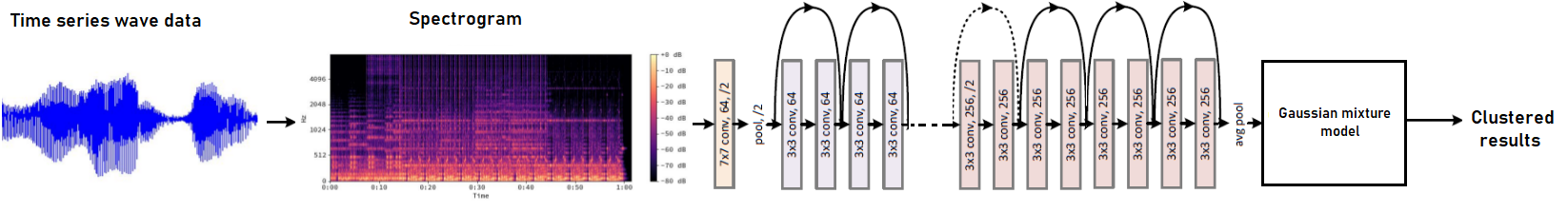}
    \caption{ Used pipeline }

\end{figure}
\begin{figure}
    \centering
    \includegraphics[scale=1.0]{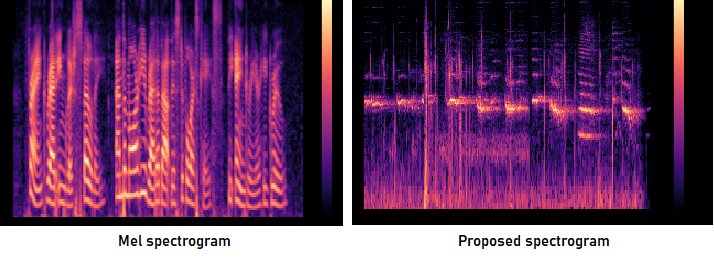}
    \caption{Mel spectrogram and the proposed spectrogram created from the same data }

\end{figure}

We used the Gaussian mixture model\cite{29} for clustering, where the goal is to find a set of K-normal distributions of mean $\mu_k$ and covariance $\sigma_k$ (where k=1 to K) to best describe the overall data. As the final output of the algorithm, a categorical variable is also inferred. The Gaussian mixture clustering is a probabilistic and more flexible version of the K-means clustering algorithm, in which the clusters can be unbalanced in terms of internal variance, each covariance can be anisotropic, and where decision boundary is soft. The negative likelihood of the data to be fully described by the set of normal distributions is used as the clustering loss. The number of clusters is inferred by our procedure. For the Gaussian mixture algorithm, we initialize the number of clusters (K) = 10 and let the model train with expectation minimization strategy\cite{29}. We made clustering optional, once clustering loss stagnated after 6000 epochs. We employed batch processing, which randomly selects subsets of the whole dataset for faster training. This also prevents the model from getting stuck in local minima. We trained the model for 10000 epochs. The Clustering loss decreased by a factor of 5.

\section{Data}
For this experiment, we used seismic data (waveforms and related metadata), publicly available on the IRIS data management center website. IRIS data services is funded by the Seismological Facilities for the Advancement of Geoscience and EarthScope (SAGE) Project, which is funded by the NSF under Cooperative Agreement EAR-1261681. In addition, we used some simulation data from USGS.

\begin{figure}
    \centering
    \includegraphics[scale=1.2]{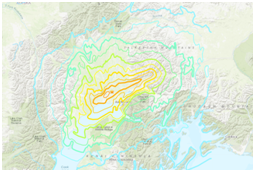}
    \caption{ M 7.5 Scenario Earthquake - Castle Mnt Earthquake2014-03-25 21:00:00 (UTC), 61.473°N 150.312°W ,10.0 km depth
Credit: U.S. Geological Survey }

\end{figure}
\begin{figure}
    \centering
    \includegraphics[scale=1.2]{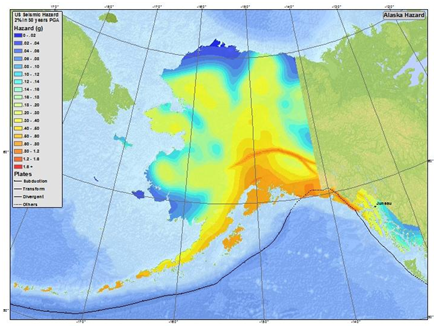}
    \caption{ 2014 Seismic Hazard Map-Alaska
Credit: U.S. Geological Survey 
Department of the Interior/USGS
}

\end{figure}

We also introduced some artificial data to the dataset. The artificial data was generated by using data augmentation techniques (translation, reflection about horizontal and vertical axis).
\section{Results}
Our method consistently outperformed other, commonly used models, with a similar number of parameters. In addition, our model was able to train well on smaller datasets. In the comparative study between the same model, trained with our variant of Mel spectrogram and ordinary spectrogram, the one trained on the variant of Mel spectrogram outperformed the one with normal spectrogram by a substantial margin. [Table \ref{table: 1}]

\begin{table}[]
\centering
\begin{tabular} {||c | c | c||}
\hline
  &                                       &                                  \\
CNN & Clustering loss with ordinary spectrogram & Clustering loss with Mel spectrogram \\ 
  &                                       &                                  \\[0.5ex]
\hline\hline
  &                                       &                                  \\

Resnet 18  & 6.27                                      & 5.04                                 \\
Resnet 50  & 5.62                                      & 4.50                                 \\
Resnet 101 & 4.79                                      & 3.95                                 \\
Resnet 152 & 3.56                                      & 2.90 \\ [1ex]

\hline
\end{tabular}
\hspace{\textwidth}
\caption{Clustering loss comparison for different depths of Resnets with ordinary spectrogram and our spectrogram}
\label{table: 1}
\end{table}

\begin{figure}
    \centering
    \includegraphics[scale=0.5]{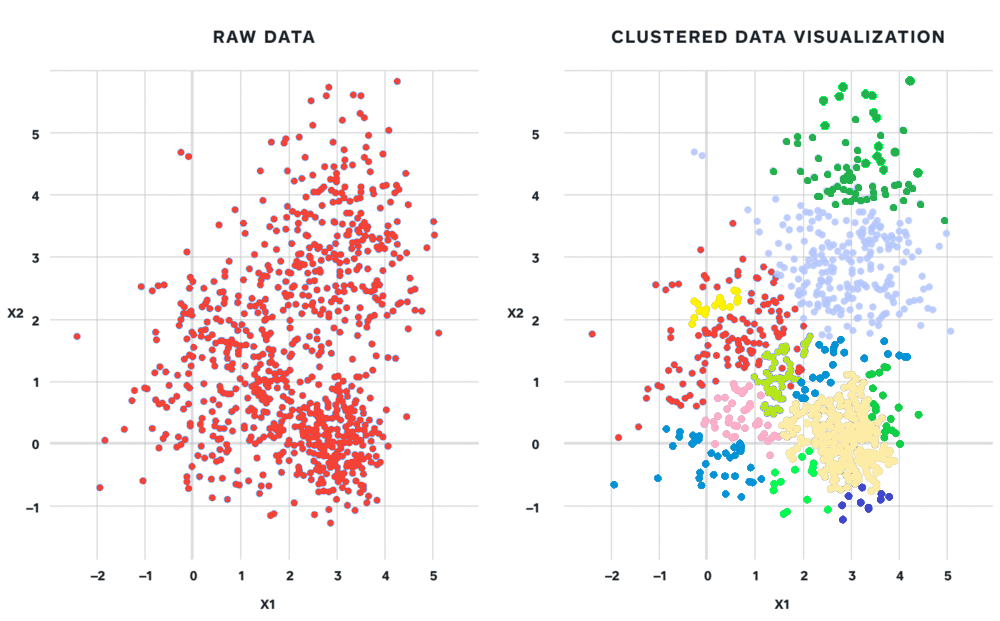}
    \caption{ Visualization of data pre and post clustering
}

\end{figure}
\section{Discussion}
Our method, thus, provides a valuable tool for remote seismic monitoring centers, who don’t have access to a high amount of computational resources. The clustering algorithms can detect very subtle patterns, which might be missed by human seismologists, and group data items together. Thus, it is particularly effective for data shown in [30,31]. The clustering also allows the detection of seismological phenomena that might have been considered regular and ignored by a seismologist. This is important for spotting precursors to a natural disaster, like an earthquake or volcanic eruption. It also enables a better understanding of the inner mechanics of the Earth’s crust by detecting hard to detect phenomena, like nonvolcanic tremors, low-frequency earthquakes, distant vs local earthquakes, etc.
Animals like elephants have been widely known to run inland from the coast just before an earthquake, followed by a tsunami (https://www.nationalgeographic.com/animals/article/news-animals-tsunami-sense-coming, https://www.dailymail.co.uk/news/article-3614477/How-Ning-Nong-elephant-saved-tsunami-incredible-bond-little-girl-baby-jumbo-inspired-Michael-Morpurgo-s-sequel-War-Horse.html). Since hearing of animals has not been empirically tested to generate proper constants for the Mel spectrogram, we tried out different constants stochastically and reported the results for the best one, so far, in this paper. We are continuing to test more combinations of constants and better results could be forthcoming.
If we get close to the hearing efficiency of an elephant then every seismological monitoring station would be able to detect an approaching earthquake and issue an early warning of up to 20 minutes before the earthquake hits, saving lives of people in the region by giving them sufficient time to get to safe places.
\section{Conclusion}
We have proposed an idea that, to our knowledge, has not been explored before. It is far from perfect as the manual testing of constants for the Mel Spectrogram variant is tedious and time-consuming. We are currently working on automating the process. There is a lot of work to be done in the field of ML for seismology, which due to lack of attention from the industry, remains largely neglected. It is a very important area of study with the potential to save thousands of lives each year.

\bibliographystyle{unsrtnat}


\end{document}